\documentclass{egpubl}
\usepackage{eg2023}
 
\ShortPresentation      % uncomment for (final) Short Conference Presentation
\usepackage[T1]{fontenc}
\usepackage{dfadobe}  

\BibtexOrBiblatex
\electronicVersion
\PrintedOrElectronic
\ifpdf \usepackage[pdftex]{graphicx} \pdfcompresslevel=9
\else \usepackage[dvips]{graphicx} \fi

\newcommand{\R}{\mathbb{R}}
\usepackage[acronym]{glossaries}
\glsdisablehyper

\newacronym[plural=PCFGs,firstplural=probabilistic context-free grammars (PCFGs)]{pcfg}{PCFG}{probabilistic context-free grammar}
\newacronym[plural=L-Systems,firstplural=Lindenmayer-Systems (L-Systems)]{lsystem}{L-System}{Lindenmayer-System}
\newacronym[plural=TLSs,firstplural=Terrestrial Laser-Scanners (TLSs)]{tls}{TLS}{Terrestrial Laser-Scanner}

\newacronym[plural=MCMCs,firstplural=Monte Carlo Markov Chain (MCMCs)]{mcmc}{MCMC}{Monte Carlo Markov Chain}
\newacronym[plural=CNNs,firstplural=Convolutional Neural Networks (CNNs)]{cnn}{CNN}{Convolutional Neural Network}
\newacronym[plural=LSTMs,firstplural=Long Short-Term Memory Networks (LSTMs)]{lstm}{LSTM}{Long Short-Term Memory Network}
\newacronym{nlp}{NLP}{natural language processing}
\newacronym{ppl}{PPL}{perplexity}
\newacronym{bpc}{BPC}{bits-per-character}

\usepackage{egweblnk}
\usepackage{amsmath}
\usepackage{amssymb}
\usepackage{multirow}
\usepackage{subcaption}
\title[Towards L-System Captioning for Tree Reconstruction]%
      {Towards L-System Captioning for Tree Reconstruction}

\author[J. Magnusson, A. Hilsmann, P. Eisert]
{\parbox{\textwidth}{
        \centering
        J.\,S. Magnusson$^{1}$\thanks{This research was grant-aided by the German Federal Ministry of Economic Affairs and Climate Action as part of the NaLamKI project under Grant 01MK21003D and the state Berlin (ProFIT) under grant number 10174498 (BerDiBa)}\orcid{0000-0002-3913-735X},
        A. Hilsmann$^{1}$\orcid{0000-0002-2086-0951} and
        P. Eisert$^{1,2}$\orcid{0000-0001-8378-4805}
    }
    \\
{\parbox{\textwidth}{
        \centering $^1$Fraunhofer Heinrich Hertz Institute HHI, Germany\\
         $^2$Visual Computing, Institut für Informatik, Humboldt-University, Germany
    }
}
}
\begin{document}

% uncomment for using teaser
\teaser{
 \includegraphics[width=0.5\linewidth]{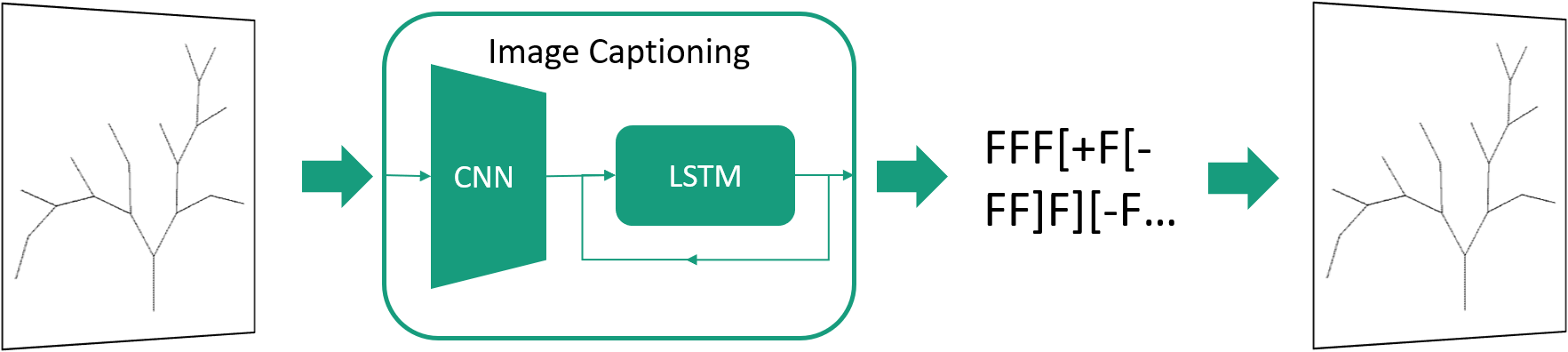}
 \centering
  \caption{Our goal is to train a model that can infer L-System words as a representation of trees from images.}
\label{fig:pipeline}
}

\maketitle
%-------------------------------------------------------------------------
\begin{abstract}
This work proposes a novel concept for tree and plant reconstruction by directly inferring a \gls{lsystem} word representation from image data in an image captioning approach. We train a model end-to-end which is able to translate given images into \gls{lsystem} words as a description of the displayed tree. To prove this concept, we demonstrate the applicability on 2D tree topologies. Transferred to real image data, this novel idea could lead to more efficient, accurate and semantically meaningful tree and plant reconstruction without using error-prone point cloud extraction, and other processes usually utilized in tree reconstruction. Furthermore, this approach bypasses the need for a predefined \gls{lsystem} grammar and enables species-specific \gls{lsystem} inference without biological knowledge.
\\
%-------------------------------------------------------------------------
%  ACM CCS 2012 (see http://www.acm.org/about/class/class/2012)
%The tool at \url{http://dl.acm.org/ccs.cfm} can be used to generate
% CCS codes.

\end{abstract}  
%-------------------------------------------------------------------------
\section{Introduction}
\label{sec:intro}

Tree modeling and reconstruction are challenging problems in Computer Graphics and Vision and have gained increased interest with the digitalization in agriculture.
Current methods for tree and plant reconstruction mostly start with point clouds measured by 3D scanners or extracted from multi-view images. They derive surface meshes or skeletons as a final representation \cite{okura_3d_2022}. Reconstruction based on photogrammetric approaches is challenging due to occlusions, complex structures, or measuring errors. In modeling, fractal structures like trees and plants are generated by procedural modeling based on grammar- or graph-based representations \cite{chaudhuri_learning_2020}. For example, \acrfullpl{lsystem} were developed to model plant topologies. \Glspl{lsystem} are formal grammars which produce a sequence of commands. These commands describe geometric shapes based on turtle geometry \cite{prusinkiewicz_algorithmic_2012}. Further extensions allow for 3D modeling, growth simulations, or even nutrient propagation inside plants which makes \glspl{lsystem} a powerful tool for biological computations \cite{prusinkiewicz_algorithmic_2012}. For these computations, \gls{lsystem} words are more suited to represent individual plants and trees than point clouds or meshes, which are usually derived in reconstruction. Therefore, we bridge the gap between the usual reconstruction and modeling processes and directly infer an \gls{lsystem} representation of a tree from image data.

The covered problem shows parallels to that in image captioning, where an image is translated into natural language describing the image content. It also relates to inverse procedural modeling which models a procedural representation of existing geometries. Inspired by advances in these areas, we propose to treat the problem as an image captioning problem to translate an image of a tree into a grammar-based representation used in inverse procedural modeling: \glspl{lsystem} words.

In this paper, we present a proof of concept on simple 2D images as an important step towards solving the problem of inverse procedural modeling of trees and plants from images. The contributions of this work are the following:

\begin{enumerate}
    \item We present a new concept that combines advances in \gls{nlp} and image captioning with procedural modeling approaches for plants to directly infer an \gls{lsystem} representation from image data.
    
    \item We proof the presented concept on simple 2D tree topology images which are automatically translated into \gls{lsystem} words. This approach is planned to be adapted to more complex data, i.e.~RGB images.
    
    \item We define requirements for an \gls{lsystem} word to ensure a bidirectional relation between image and derivation.
\end{enumerate}

This paper is organized as follows. Sec.~\ref{sec:relwork} presents related work before Sec.~\ref{sec:infer} describes our proposed approach. Sec.~\ref{sec:method} presents the dataset and experimental setup. Then Sec.~\ref{sec:eval} evaluates our method for \gls{lsystem} captioning from images using standard \gls{nlp} metrics. 

%---------------------------------------------------------------
\section{Related Work}
\label{sec:relwork}
While reconstruction refers to the process of extracting a precise 3D representation of a specific tree, modeling refers to approaches that are able to synthesize a representation of a virtual tree \cite{okura_3d_2022}. Most approaches that use image data for tree reconstruction first extract 3D point clouds using photogrammetric procedures \cite{guo_realistic_2020, gao_novel_2021} or start with measured point clouds by e.g. Terrestrial Laser Scanners \cite{liu_treepartnet_2021, zhen_wang_structure-aware_2014} from which specific representations of shapes and structures are extracted \cite{okura_3d_2022}.
The main challenges of using point clouds is their noisiness and overlapping structures where only the surface of the object was captured.
As the final representation, most works on image-based tree and plant reconstruction use surface meshes \cite{zhen_wang_structure-aware_2014, guo_realistic_2020}.

Regarding tree modeling, procedural modeling approaches are usually exploited when modeling trees \cite{chaudhuri_learning_2020}. Guo et al.~\cite{guo_realistic_2020} estimate parameters like angles, diameter transmission coefficients, and the number of segment lengths from a point cloud. These parameters guide a procedural model to grow a tree filling the input point cloud. Similarly, Stava et al.~\cite{stava_inverse_2014} estimate 24 parameters via a \acrlong{mcmc} and a similarity measure incorporating shape, geometry, and structure similarities. As an input they use a tree graph representation. Guo et al.~\cite{guo_inverse_2020} detect atomic structures of tree topologies with an RCNN to infer a compact \gls{lsystem} grammar. This method is also capable of reconstructing user drawn sketches of 2D topologies. However, an adaptation to more complex data and 3D structure is difficult as it depends on detectable structures in an image. In contrast, we are interested in developing a modular method adaptable to real image data, that learns to reconstruct topologies even if they are invisible.

Our goal is to infer an \glspl{lsystem} representation directly from an image. The task is similar to image captioning problems where images are translated into text using deep learning approaches. Typically, information in images is extracted and decoded into natural language. Mostly, a \gls{cnn} is used as an encoder and a recurrent network or Transformer decodes the information extracted by the encoder \cite{stefanini_show_2021}. The image can be encoded globally \cite{wu_image_2018, donahue_long-term_2015} or separately over a grid \cite{lu_knowing_2017} or visual regions. The encoding is then aggregated by an attention mechanism. The decoding process was mostly done by \glspl{lstm} \cite{wu_image_2018, donahue_long-term_2015, lu_knowing_2017, anderson_bottom-up_2018}. Lately, more methods focus on Transformers due to their advances in \gls{nlp} \cite{li_oscar_2020}.

For this work, we focus on the most simplistic approach to prove the applicability of image captioning to the problem of reconstructing trees. Additionally, instead of inferring a procedural model and its parameters for a given input, an \gls{lsystem} word is directly derived with image captioning. The resulting word is the final representation instead of surface meshes or point clouds usually used in reconstruction pipelines. From multiple derivations, a generalized \gls{lsystem} can be easily inferred. This enables biological computations such as growth simulations or nutrient propagation which can be realized by \glspl{lsystem}.

%-----------------------------------------------------------------
\section{Inferring L-System Words}
\label{sec:infer}
The proposed method receives an image of a tree topology. This image is transcribed into an \gls{lsystem} word via an established image captioning pipeline with a \gls{cnn}+\gls{lstm} backbone (Fig. \ref{fig:pipeline}). In this work, we use an LSTM-based architecture instead of e.g. more state-of-the-art Transformers to demonstrate that simple methods are capable of solving this problem. As a proof of concept, we focus on 2D tree topologies generated by a probabilistic \gls{lsystem}. Before describing the method in detail, we formally define \glspl{lsystem} and heuristic rules to ensure a bidirectional mapping between words and images.

\begin{figure}
  \centering
  \includegraphics[width=.8\linewidth]{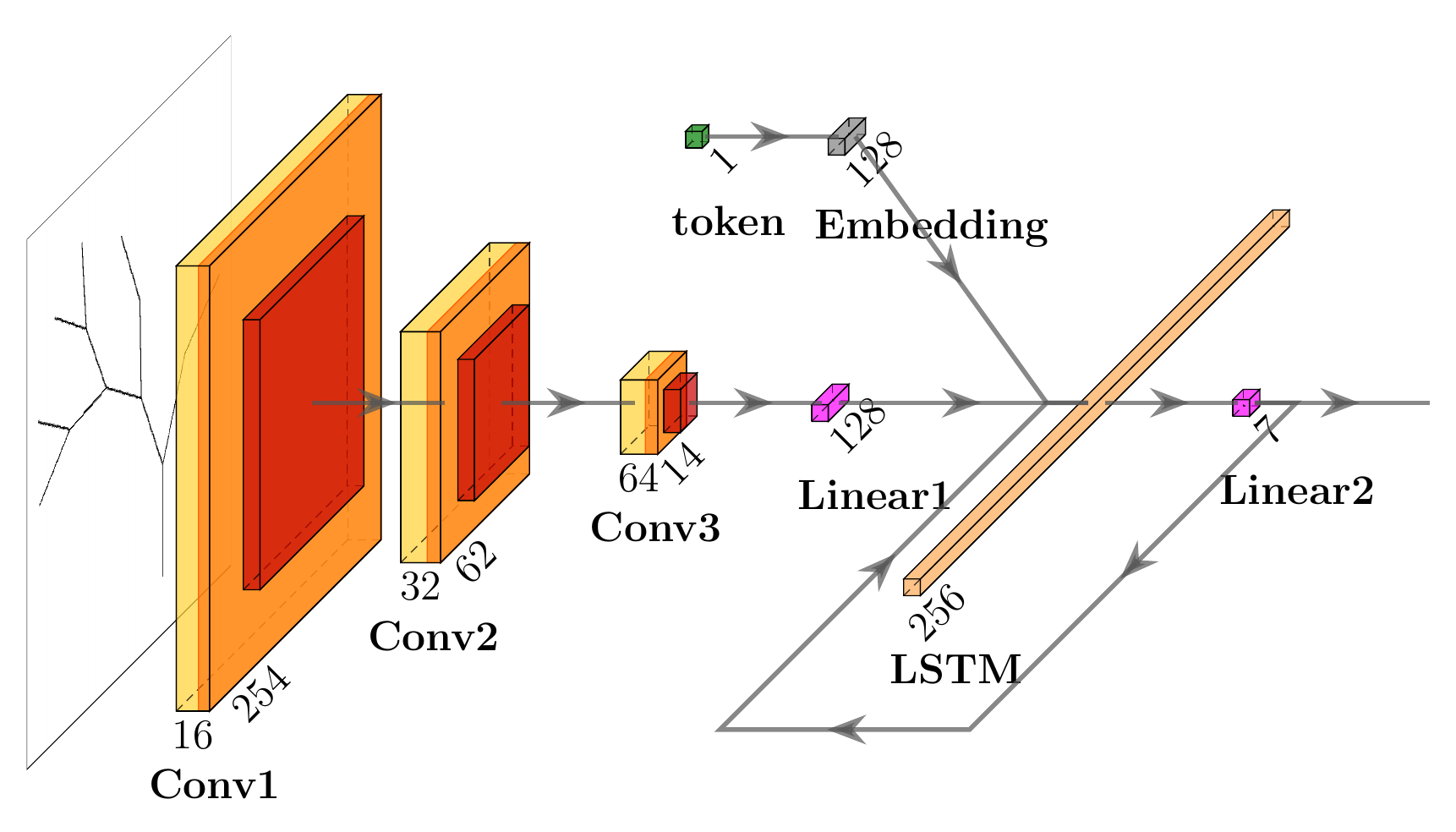}
  \caption{\label{fig:arch}Model to reconstruct \gls{lsystem} words from images. Each convolution is followed by ReLUs. Red displays max pooling.}
\end{figure}

\subsection{Lindenmayer Systems}
\label{sec:lsystems}
Let $V$ be an alphabet with its set of words $V^*$ and nonempty words $V^+$. We define \glspl{lsystem} as a grammar $G = (V, \omega, P, \pi, \delta, f)$ with the axiom $\omega \in V^+$, a set of productions $P \subset V \times V^*$, an branching angle $\delta \in \R$ and a F-length $f \in \R^+$. A production is interpreted as the substitution of a symbol $a \in V^+$ with a word $b \in V^*$ \cite{prusinkiewicz_algorithmic_2012}. For probabilistic \glspl{lsystem}, $\pi: P \to (0,1]$ defines a mapping of productions to an occurrence probability. $\delta$ and $f$ are constants that are required to display an \gls{lsystem} word.

For the 2D case, the alphabet is defined as $V = {F, +, -, [, ]}$. Each symbol can be interpreted as a command for a turtle which navigates on a plane and draws its path (see \ref{tab:lsystem-symbols}).

\begin{table}
  \centering
  \begin{tabular}{cc}
    Command & Description \\
    \hline
    $F$ & move forward and draw a line \\
    $+$ & rotate right \\
    $-$ & rotate left \\
    $[$ & save current state \\
    $]$ & restore last saved state \\
  \end{tabular}
  \caption{\gls{lsystem} commands to draw turtle geometry.}
  \label{tab:lsystem-symbols}
\end{table}

In order to build a grammar which can describe structures in an image, we ensure a bidirectional relationship between image and \gls{lsystem} word. We identified five ambiguities regarding the relation of image and \gls{lsystem} word. From these we derived five rules which must be fulfilled by all \gls{lsystem} words to ensure a bi-direction between both modalities. By these, we process and filter our generated data to stabilize the training procedure.

\begin{enumerate}
 \item An \gls{lsystem} word might contain segments that not only cross each other but be directly on top of each other: We exclude these cases by calculating the positions of all segments and filter out doubled segments.
 \item Rotations can cancel each other out: +/- cannot follow -/+
 \item Branches can be empty and therefore invisible: forbid [ ]
 \item Branches starting at the same point can be ordered differently: we sort branches from right to left, e.g. F[-F][+F]F is rewritten as F[+F][-F]F
 \item The last segment of a branch can be a branch or the extension of a previous branch: a branch cannot end with a subbranch, e.g. F[+F] must be rewritten as F+F
\end{enumerate}

%---------------------------------------------------------------
\subsection{Learning L-System Image Captions}
Our model follows a standard image captioning pipeline with a \gls{cnn} that encodes global features from the input image. The flattened encoding is resized by a linear layer and concatenated to each embedded token. This vector is inserted into a 1-layer \gls{lstm} to predict the next token until a maximal sequence length is reached or the \textless eos\textgreater~token is predicted (Fig.~\ref{fig:arch}). Tokenization describes the process of splitting a string into its entities defined by the vocabulary e.g. using its characters or words. Regarding the tokenization for \glspl{lsystem}, we utilize the vocabulary $\{$\textless bos\textgreater, \textless eos\textgreater$\}$ + $\{$F, +F, -F, [, ]$\}$ (compare with Tab.~\ref{tab:lsystem-symbols}). Because each rotation is always followed by an F, we fuse these to the tokens +F and -F to prevent violations of rule 2 from Sec. \ref{sec:lsystems}. In the evaluation, we show that this strategy is advantageous over character based tokenization. Moreover, rotations can be parameterized so that multiple rotations are unnecessary to enable different rotations per branch.

%---------------------------------------------------------------
\section{Experimental Setup}
\label{sec:method}

\begin{figure}
    \centering
    \includegraphics[width=0.8\linewidth]{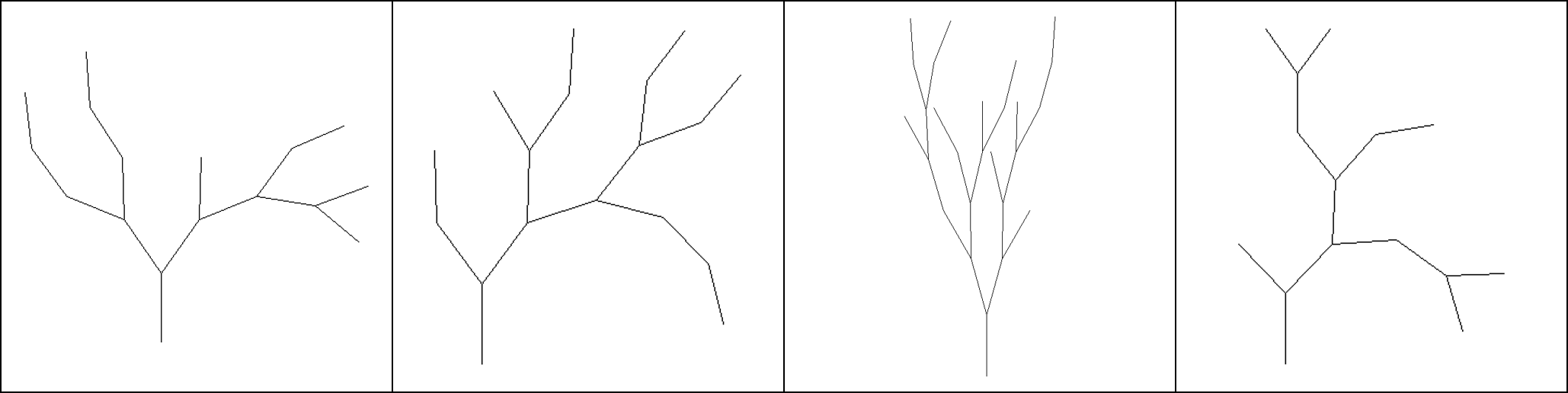}
    \caption{Example images with a resolution of $512 \times 512$ px.}
    \label{fig:example-inputs}
\end{figure}

For the training data, we generated 48267 unique samples from a simple probabilistic \gls{lsystem} with a random number of derivations $n \in [1, 7]$ which follow the rules from Sec. \ref{sec:lsystems}. For data augmentation, the angle $\delta$ is randomly set to $\delta \in [15^{\circ}, 60^{\circ}]$ each epoch while the F-length $f = 100$ stays constant. The angle interval is chosen arbitrarily based on example values in \cite{prusinkiewicz_algorithmic_2012}. The tree is rendered so that the tree spans the whole image. No further augmentation is applied since the data is not noisy. We apply a 0.9, 0.05, 0.05 split for training, validation, and testing. Examples of the input images are given in Fig.~\ref{fig:example-inputs}.

The proposed model is trained for 495 epochs with a learning rate of 0.00025 and an ADAM optimizer. Without teacher forcing, the model optimizes the cross-entropy loss of all predicted tokens. Therefore, the loss is equally sensitive to all tokens and enforces the exact \gls{lsystem} word given by the ground truth.

\section{Evaluation}
\label{sec:eval}
% 0.1214 in epoch 486
For numerical evaluation, we consider standard \gls{nlp} metrics like \gls{bpc} and \gls{ppl} and analyze the correctness in different categories such as correct, false syntax, non-terminated, and residue. Correct and non-terminating sequences are self-explanatory. Any syntax error or violation of the rules from Sec.~\ref{sec:lsystems} is treated as false syntax although some failures are non-syntactical errors. The residue category contains all words that differ in one or more tokens from the ground truth while being syntactically correct. We achieve a \gls{ppl} of 1.129 and a \gls{bpc} of 0.403 which are both proportional to the cross-entropy of 0.1214 for the test set. A direct comparison to \gls{nlp} methods is difficult due to the usage of a distinct vocabulary and language. The categorical evaluation in Tab.~\ref{tab:categorical-eval} states that roughly 80\% of all words are reconstructed correctly.

%%%%%%%%%%%%%% rand_normal %%%%%%%%%%%%%%%%
\begin{table}[b]
    \centering
    \begin{tabular}{l|c|c|c|c}
        \multirow{2}{*}{Set} & \multirow{2}{*}{Correct} & False & Non- & \multirow{2}{*}{Residue} \\
                   &         & Syntax       & Terminating     &         \\
        \hline
        Train      & 86.23\% & 5.33\%       & 0.38\%          &  8.06\%  \\
        Validation & 80.32\% & 7.38\%       & 0.54\%          & 11.77\% \\
        Test       & 80.16\% & 7.75\%       & 0.62\%          & 11.48\% \\
        \hline
        Test Char  & 77.13\% & 8.74\%       & 0.25\%          & 13.88\% \\
        % Test-Single&         &              &                 &        
    \end{tabular}
    \caption{Categorical evaluation of the model's reconstructions. A false syntax denotes any violation of the rules from Sec.~\ref{sec:lsystems} or syntax errors. The Char numbers report the performance for testing a model using a character tokenization.}
    \label{tab:categorical-eval}
\end{table}

\begin{figure}
    \centering
    \includegraphics[width=0.9\linewidth]{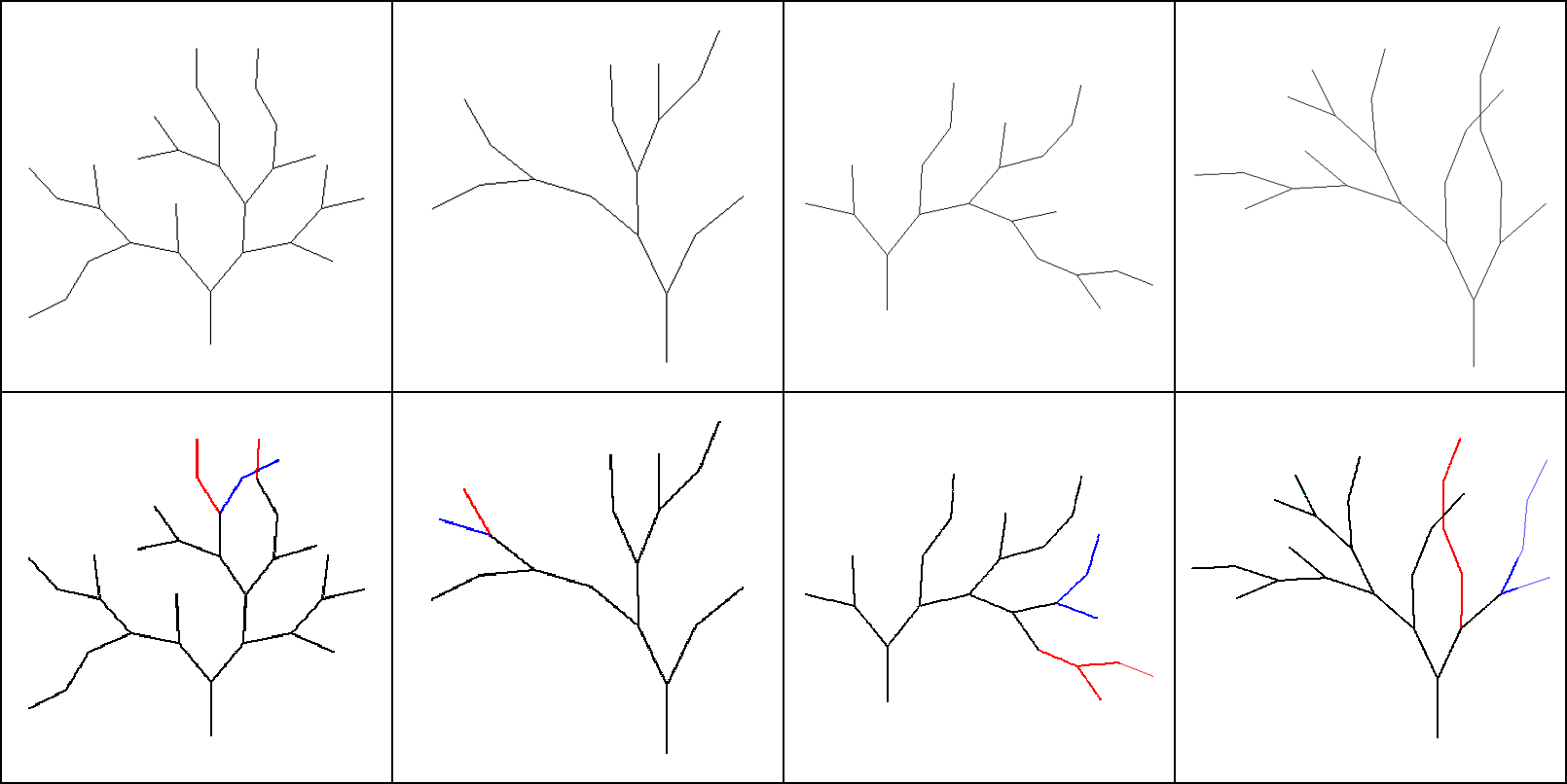}
    \caption{Examples of rendered outputs for the category residue. The top row displays the groundtruth. The bottom row shows the correct reconstructed part in black. Red and blue lines denote the part only contained in the ground truth or prediction, respectively.}
    \label{fig:example-outputs}
\end{figure}

Fig.~\ref{fig:example-outputs}, shows examples of the rendered outputs of the category residue. The most dominant errors in the residue class are wrong rotations or longer branches than given in the image. Due to the rules in Sec.~\ref{sec:lsystems}, the zeroth hierarchy is the rightmost path from trunk to leaf. The next level describes all branches originating from the previous one. The proposed method is more confident in the first levels than the character based approach (for the second level 88.54\% and 85.8\%). In training, each token and therefore hierarchy level is treated equally. Accordingly, future methods will focus on assigning more importance to lower levels which have the highest visual impact.

\label{sec:ablation}
To further validate our method, we test the effect of using a character tokenization. We argued that each rotation symbol is followed by an F and could be treated as a single token. Contrarily, one could use a character tokenization to have a completely disjoint vocabulary. Tab.~\ref{tab:fused} shows that with character tokenization violations of rule 2 are introduced. Additionally, Tab.~\ref{tab:categorical-eval} shows that the overall performance of correct \gls{lsystem} words is reduced by 3\%. To calculate the relative errors of invalid rotation groups, all invalid groups of rotation symbols are counted and divided by the total number of rotation groups. The error of invalid brackets reflects the amount of uneven opened and closed brackets to all indicated bracket pairs. Lastly, any empty bracket pair or empty branch is related to all indicated bracket pairs.

\begin{table}
    \centering
    \begin{tabular}{l|c|c|c}
         Tokenization & +- or -+ & [[]      & empty branches \\
         \hline
         Char         & 0.17 \%  &  1.32 \% & 0.05 \% \\
         Fused        & -        &  1.51 \% & 0.19 \% \\
    \end{tabular}
    \caption{Errors in the test set for character (Char) and rotation-fused (Fused) tokenization. The first error describes a violation of rule 2 from Sec.~\ref{sec:lsystems}. $[[]$ refers to invalid bracket pairs.}
    \label{tab:fused}
\end{table}

%-----------------------------------------------------------------
\section{Conclusion}
\label{sec:conclusion}
This paper presents a novel approach to reconstruct \gls{lsystem} word representations for trees from images. It combines ideas from procedural modeling, image captioning, and \gls{nlp}. Using a simple \gls{cnn}+\gls{lstm} architecture, we translate images of tree topologies into corresponding \gls{lsystem} words. Using this representation, we bypass the necessity of a predefined \gls{lsystem} grammar, used in procedural modeling, or the explicit object detection used by Guo et al.~\cite{guo_inverse_2020}.
Furthermore, our approach allows to infer a species-specific \gls{lsystem} from a collection of words without biological and phenological knowledge using grammar inference techniques. These grammars enable biological computations such as growth modeling.

We proof this concept with simplistic data and achieve promising results that underline the applicability of image captioning to reconstruct trees using the representation of \gls{lsystem} words. Future work will concentrate on enabling different rotation angles and segment lengths inside 
 a tree. Furthermore, we will incorporate multiview constraints and synthetic 3D data in order to advance the current approach towards more challenging and real data. Once this is achieved, we will overcome challenges of point cloud reconstruction for trees.

\bibliographystyle{eg-alpha}

\bibliography{references_reduced}

\end{document}